\documentclass[conference]{IEEEtran}
\IEEEoverridecommandlockouts
\usepackage{cite}
\usepackage{amsmath,amssymb,amsfonts}
\usepackage{algorithmic}
\usepackage{graphicx}
\usepackage{textcomp}
\usepackage{xcolor}
\usepackage{multirow}
\usepackage{booktabs}
\def\BibTeX{{\rm B\kern-.05em{\sc i\kern-.025em b}\kern-.08em
    T\kern-.1667em\lower.7ex\hbox{E}\kern-.125emX}}
\begin{document}

\title{A Data-Centric Approach for Improving \\ Adversarial Training Through the Lens of  Out-of-Distribution Detection
\thanks{\textsuperscript{1}Equal contribution.}
}

\author{\IEEEauthorblockN{Mohammad Azizmalayeri\textsuperscript{1}}
\IEEEauthorblockA{\textit{Computer Engineering Department} \\
\textit{Sharif University of Technology}\\
Tehran, Iran \\
m.azizmalayeri@sharif.edu}
\and
\IEEEauthorblockN{Arman Zarei\textsuperscript{1}}
\IEEEauthorblockA{\textit{Computer Engineering Department} \\
\textit{Sharif University of Technology}\\
Tehran, Iran \\
arman.zarei@sharif.edu}
\and
\IEEEauthorblockN{Alireza Isavand}
\IEEEauthorblockA{\textit{Computer Engineering Department} \\
\textit{Sharif University of Technology}\\
Tehran, Iran \\
alireza.isavand@sharif.edu}
\and
\IEEEauthorblockN{\ \ \ \ \ \ \ \ \ \ \ \ \ \ \ \ \ \ \ \ \ \ \ \ 
Mohammad Taghi Manzuri}
\IEEEauthorblockA{\textit{\ \ \ \ \ \ \ \ \ \ \ \ \ \ \ \ \ \ \ \ \ \ \ \ 
Computer Engineering Department} \\
\textit{\ \ \ \ \ \ \ \ \ \ \ \ \ \ \ \ \ \ \ \ \ \ \ \ 
Sharif University of Technology}\\
\ \ \ \ \ \ \ \ \ \ \ \ \ \ \ \ \ \ \ \ \ \ \ \ 
Tehran, Iran \\
\ \ \ \ \ \ \ \ \ \ \ \ \ \ \ \ \ \ \ \ \ \ \ \ 
manzuri@sharif.edu}
\and
\IEEEauthorblockN{Mohammad Hossein Rohban}
\IEEEauthorblockA{\textit{Computer Engineering Department} \\
\textit{Sharif University of Technology}\\
Tehran, Iran \\
rohban@sharif.edu}
}

\maketitle

\begin{abstract}
Current machine learning models achieve super-human performance in many real-world applications. 
Still, they are susceptible against imperceptible adversarial perturbations.
The most effective solution for this problem is adversarial training that trains the model with adversarially perturbed samples instead of original ones. 
Various methods have been developed over recent years to improve adversarial training such as data augmentation or modifying training attacks. 
In this work, we examine the same problem from a new data-centric perspective. 
For this purpose, we first demonstrate that the existing model-based methods can be equivalent to applying smaller perturbation or optimization weights to the hard training examples. 
By using this finding, we propose detecting and removing these hard samples directly from the training procedure rather than applying complicated algorithms to mitigate their effects. 
For detection, we use maximum softmax probability as an effective method in out-of-distribution detection since we can consider the hard samples as the out-of-distribution samples for the whole data distribution. 
Our results on SVHN and CIFAR-10 datasets show the effectiveness of this method in improving the adversarial training without adding too much computational cost. 

\end{abstract}


\begin{figure*}[t]
\centerline{\includegraphics[scale=0.3]{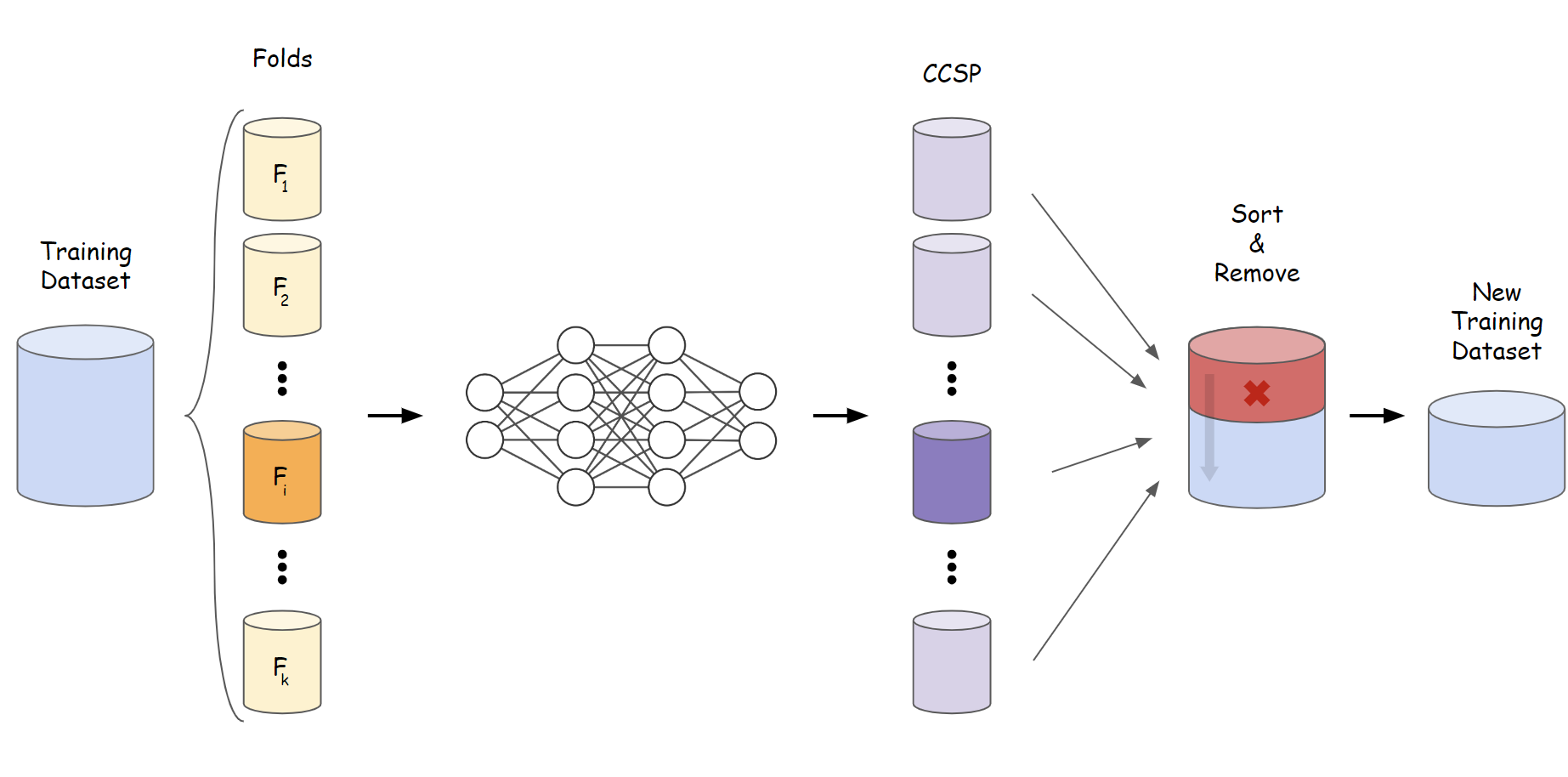}}
\caption{Overview of our approach. The training set is first divided into $k$ folds. After that, each time, we train model using $k-1$ of folds, and the CCSP scores (our measure to detect hard training samples) are calculated for samples of remaining fold using the trained model. By repeating this for all folds, the CCSP score would be calculated for all training samples using this method. Then, we sort scores, and remove $R$ samples with the lowest CCSP scores. Finally, the model is trained on this new purified dataset rather than the original dataset.}

\label{fig_1}
\end{figure*}

\section{Introduction}

In recent years, deep neural networks (DNNs) are proving to be successful in a variety of applications, such as image processing \cite{vision}, Natural Language Processing \cite{NLP1}, etc. Nevertheless, we cannot still rely on them since they are subject to adversarial examples that cannot even be recognized by humans \cite{AT1, AT2, AT3}. Adversarial examples are generated by adding an optimized $\ell_p$ norm-bounded perturbation to the original samples. Perturbing a sample within an $\epsilon$-ball around it may change the prediction and significantly impact the model's performance while in many applications, such as autonomous driving \cite{auto}, robustness of the models against these attacks is critical.

In order to achieve robust models against adversarial attacks, several approaches have been proposed, with Adversarial Training (AT) \cite{AT} proving to be the most effective. The purpose of this method is to learn a robust model through solving a min-max problem. Briefly, AT first tries to find the perturbations within the $\epsilon$-ball that causes the maximum loss for each perturbed sample, which is referred as the maximization part. Next, the model loss is minimized on the perturbed samples rather than the original ones to learn more robust features against the adversarial perturbations, which is referred as minimization part. The minimization is done with gradient descent method, while the maximization is often done using an attack called Projected  Gradient Descent (PGD) \cite{AT}. AT can be formulated as:
\begin{equation}\label{my_first_eqn}
\min_{\theta} \frac{1}{n} \sum_{i=1}^{n} \max_{{\left \| x'_i-x_i \right \|}_\infty\leq\epsilon}\mathcal{L}(f_\theta(x'_i),y_i)
\end{equation}
where $\mathcal{L}(f_\theta(x'_i),y_i)$ is the prediction loss of model $f_\theta$ with parameters $\theta$ on the perturbed sample $(x'_i, y_i)$ within an $\epsilon$-ball around $x_i$.

Recently, several approaches have been proposed to enhance adversarial training. For instance, data augmentation methods are used to improve robustness by applying better augmentations or adding generated data in the training \cite{Rebuffi2021}, or perceptual training is suggested to cover a wider range of training perturbation types \cite{perceptual}.

In contrast with these methods, there are some other approaches that aim at differentiating between different samples by changing the optimization weight in the training or the training perturbation budget for each sample, without changing the other training settings.
For instance, \cite{balaji2019} improves robustness by setting a sample-based perturbation budget during adversarial training to move all the training samples to the model's decision boundaries during training.

Following these methods, we want to improve robustness by differentiating between the samples with a new data-centric approach. To this end, we try to modify the original training set itself, but not by using training augmentation techniques, or changing the training attack methods, or etc. In other words, our efforts in this work is to fix the data which the code is running on instead of changing the training method, which can be classified as a data-centric approach.

For this purpose, we first note that the existing methods that differentiate between different samples mainly try to reduce the effect of the samples in decision boundaries in the training process as discussed in section \ref{sec_method_1}. Accordingly,
some samples in the training set are hard to learn for the model during training. It can happen when samples are near boundaries, or outside of their classes distributions due to the reasons such as wrong labeling. Forcing the model to learn these samples can reduce the generalization ability of the model to the test samples. This problem gets even worse in AT since an $\epsilon$-ball is enforced around those hard samples.

To mitigate this issue with a data-centric approach, we propose to identify these samples, and improve the data quality by deleting them from the dataset, regardless of the training method and the setup. These samples can be detected before the training process, which we call it the ``offline" method, or adaptively during training, which we call the  ``online" method.
Moreover, for identifying these samples, we utilize softmax probability as a measure that can determine whether a sample is out-of-distribution \cite{MSP}. Finally, the remaining samples are used to train the model after the hard samples are detected and removed from the training dataset. This method is shown schematically in Fig. \ref{fig_1}.

Our Results on the CIFAR-10 and SVHN datasets demonstrate that this data-centric strategy can enhance model's robustness with both ``offline" and ``online" methods, without significantly raising computational cost.  We also point out that softmax probability as the detection method can be substituted with Mahalanobis distance \cite{Mahalanobis}, and the findings still show improvement. We hope this work to be a start on using data-centric approaches in adversarial training.

\section{Related Works}
Deep networks are vulnerable to attacks, while defenses attempt to achieve a robust model against them. In the following, some popular adversarial attacks and defenses are explored.

\subsection{Attack}
The threat of adversarial examples was first noticed in image classification models \cite{AT2}. The accuracy of the model can be significantly reduced when a norm-bounded perturbation is added to the input. This perturbation can be generated with iterative updates based on the loss function gradient \cite{Kurakin2018} as:
\begin{equation}
    \delta_t = \delta_{t-1} + \alpha . sign (\nabla_x J(\theta, x, y)),
\end{equation}
where $\delta_t$ is the perturbation at step $t$, and $J(\theta, x, y))$ is the cost used to train the neural network with parameters $\theta$. To ensure that the perturbation is imperceptible, it can be projected to the $\ell_p$-norm ball which is known as the Projected Gradient Descent (PGD) attack \cite{minmax}. There are also some other powerful attacks, but PGD is regarded as a standard attack for training and evaluation.

\begin{figure*}[t]
\centerline{\includegraphics[scale=0.35]{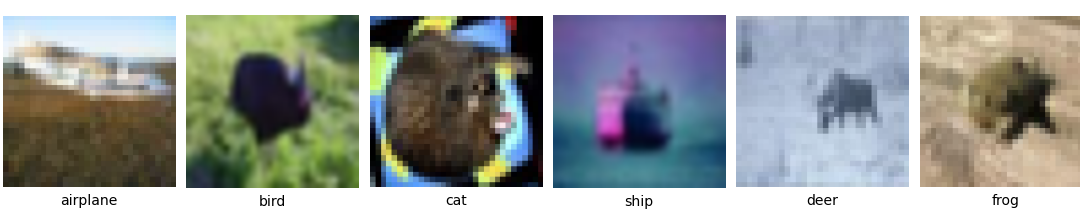}}
\caption{Some images from CIFAR-10 dataset whose labels are hard to recognize even by humans.}
\label{fig_2}
\end{figure*}

\subsection{Defense}\label{sec_defense}
Several methods have tried to stand against adversarial examples, but they mostly give a false sense of robustness due to the reasons such as gradient obfuscation or vulnerability against newer attacks \cite{AT}. Still, the best existing defense is adversarial training \cite{minmax}, which trains the model with adversarial examples. Due to the effectiveness of  adversarial training, recent methods have tried to improve it by methods such as data augmentation \cite{Rebuffi2021}, training attack modification \cite{Azizmalayeri2021}, and model hyper-parameters tuning \cite{pang2021}.

\section{Method}

\subsection{A new perspective on some of existing defenses}\label{sec_method_1}
As mentioned in section \ref{sec_defense}, there are different variants of adversarial training. Taking a closer look at them from a different angle, we can demonstrate that some of them are equivalent to applying smaller perturbation or optimization weights to the hard training examples in the model optimization. To this end, a number of methods are investigated in the following.

\textbf{CAT} \cite{cat2022} and \textbf{IAAT} \cite{balaji2019}: These methods hypothesize that the poor generalization of adversarial training is a consequence of uniform perturbation radius around every training sample. They suggest a sample-specific perturbation radius that increases the radius until the perturbed sample is miss-classified. Thus, the hard training examples would have smaller perturbation budgets during training to simplify the training of the model.

\textbf{MART} \cite{MART2020}: This method proposed to explicitly differentiate the misclassified and correctly classified examples during the training by setting lower training optimization weight for the misclassified unperturbed samples. Misclassified samples are presumably near the boundary, so they can be considered hard training examples for the model, which this method suggests assigning a lower optimization weight to.

\textbf{TRADES} \cite{trades2019}: There is a trade-off between clean and adversarial accuracy. To mitigate this issue, TRADES suggests to trade adversarial robustness off against accuracy by training model on clean examples while minimizing the KL divergence of model prediction for clean and perturbed samples. Hence, if a sample is near boundary and have a high prediction error, the KL divergence term would get a smaller value and the model would focus on learning the clean example.

\textbf{Early stopping} \cite{overfitting2020}: It is shown that overfitting to the training set harms robust performance in adversarially robust training. In other words, model can learn adversarially perturbed training samples with a high accuracy but it does not generalize well to the test samples. As a solution, the training can be stopped early if overfitting occurs using a validation set. This is equivalent to stop training before learning out-of-distribution or hard samples.

\textbf{Noisy label} \cite{Zhu2021}: Similar to standard training, adversarial training also suffers from noisy labels during training. A measure that can help to detect the noisy samples is the number of PGD iterations needed to generate misclassified adversarial examples. In other words, this measure also tries to find the near-boundary training examples using the number of PGD steps as a distance measure.

\subsection{Proposed Method}

When training a model, we might come across a number of samples that are close to the decision boundaries or even outside of the training classes distributions. 
These samples might have been produced as a result of a labeling error, or other reasons such as conceptual ambiguity in the images that makes classifying them even difficult for humans. Examples of such samples can be seen in the Fig. \ref{fig_2} that represents some images from the CIFAR-10 dataset.

Due to the difficulty of training a model on these samples, we refer to them as \textbf{Hard Training Samples (HTS)}. Additionally, learning an $\epsilon$-ball around these samples makes training even more challenging and raises the risk of problems like overfitting or mixed-up decision boundaries. Therefore, we believe that they may be better avoided during training as discussed in previous section on the existing methods.

There are many techniques for identifying such near-boundary or out-of-distribution samples. Detecting samples that do not fit into the classes of the training dataset is mainly known as out-of-distribution (OOD) detection in the literature. 
A simple but effective existing method for OOD detection is Maximum Softmax Probability (MSP) \cite{MSP}. This method uses $max_{c\in\{1, 2, \ldots, k\}} f_c(x)$ as the score function that classifier $f$ trained on a $k$-class dataset returns for input $x$ to be classified as the in-distribution sample.

Inspired by MSP, we propose Correct Class Softmax Probability (CCSP) as our method to detect HTS samples. CCSP for sample $x$ with label $y$ is defined as:
\begin{equation}
    CCSP(x, y) = softmax(z_y) = \frac{e^{z_y}}{\sum_j{e^{z_j}}},
\end{equation}
where $z$ is the output of classifier for the input $x$. CCSP score can be used as a measure to find the hard samples in comparison with other ones. This method has no computational overhead during the training, and can be used easily.

Now that we are able to recognize HTS, we can improve the training process by removing these samples before the training procedure begins. To this end, we need to measure the CCSP score for each sample, which is done through a $k$-fold cross-validation. Accordingly, we divide 
training dataset into $k$ folds called ${F_1, ..., F_k}$. Each time, a single fold $F_i$ is put aside and the rest are considered as the training dataset for the model to be trained. Afterward, the CCSP is calculated for the samples in $F_i$ fold. Eventually, the CCSP will have been calculated for all of the samples in the training dataset and $R$ (hyper-parameter) samples with the lowest CSSP score can be removed from the dataset. Finally, we can re-train our model using the new dataset that is resulted after removing such samples. This algorithm is called ``offline" modification of samples since all the process is done before the training starts.

Due to the changes in decision boundaries during the training process, samples that make a model's optimization struggling are not constant during training. In this line of thought, we also propose the ``online" version of our method. In this version, after each epoch in the training loop, the CCSP scores are calculated for all of the samples in the training dataset and $R$ samples with the lowest scores are removed from the training dataset just for the subsequent epoch. Note that the CCSP scores are calculated for all the training samples in each epoch, and we do not leave out the samples that were eliminated in the previous epochs.

It can be vividly understood that the offline version requires more time to be accomplished, while this has noticeably been diminished in the online version since the CCSP scores are calculated as a apart of the training process rather than the k-fold trainings prior to the main training procedure.

\section{Experiments}
We conduct experiments to demonstrate how our strategy improves model robustness. We also attempt to assess effectiveness of our method while employing online or offline detection of hard samples. In addition, for the purpose of an ablation study, we use Mahalanobis distance rather than softmax probability to identify hard samples.
We also design an experiment in the last part to show that model is capable of learning OOD samples, and conclude that these samples should be avoided in the training.


\subsection{Experimental Setup}
Two different datasets are used for evaluations, which are CIFAR-10 \cite{cifar10} and SVHN \cite{svhn}. We train our model on CIFAR-10 for 200 epochs while train it on SVHN for 100 epochs since it converges faster. Also, PreActResNet18 is used as the base model. In the training, the initial learning rate is set to 0.1 and it is multiplied by 0.1 after 50\% and 75\% of epochs. Moreover, SGD with momentum=$0.9$ and weight decay=$5e-4$ is used for the optimization.

In adversarial training, standard PGD attack with 10 iterations and a single restart is used to generate the perturbations. The perturbation is initialized randomly in the range $[-\epsilon, \epsilon]$, and bounded in an $\ell_\infty$-ball with $\epsilon=\frac{8}{255}$. Also, attack step size is set according to $\alpha = 2.5*\frac{\epsilon}{N}$, where $N$ is the number of iterations.

\subsection{Offline}
According to our ``offline" proposed method, first, the dataset is divided into four folds. Each time, one fold is selected and the model is trained on three other folds. Then, the CCSP score is calculated for each sample in the selected fold using trained model. It is done once for each fold to obtain CCSP score for all samples. After that, samples are sorted based on their CCSP scores, and then, $R$ samples with the lowest CCSP scores are eliminated from the dataset. Please note that setting $R$ to zero is equivalent to using the base adversarial training, which is our baseline in this work. 

Results are shown for different values of $R$ in Table \ref{tab_offline}. As we see, robustness is increased by removing the useless samples. On the other hand, we also note that setting $R$ to a large value can remove the useful samples in addition to the hard samples. The best performance is achieved by setting $R$ equal to 500 in CIFAR-10 and 300 in SVHN.

Please note that the results are stable and reproducible. For instance, in multiple runs of the experiment where $R=0$ samples were removed (normal adversarial training), the standard deviation of clean and robust accuracy for the CIFAR-10 dataset is $0.16\%$ and $0.54\%$, respectively.

\begin{table}[t]
\caption{Clean and robust accuracy of models trained on CIFAR-10 and SVHN datasets using ``offline" method with different number of deleted samples (R).}
  \renewcommand{\arraystretch}{1.1} 
\begin{center}
\begin{tabular}{ccccc}
\toprule
\multirow{3}{*}{$R$} & \multicolumn{4}{c}{Dataset}                                                                    \\ \cline{2-5} 
                                       & \multicolumn{2}{c|}{CIFAR-10}                             & \multicolumn{2}{c}{SVHN}            \\ \cline{2-5} 
                                       & Clean            & \multicolumn{1}{c|}{Robust}           & Clean            & Robust           \\ \hline
\multicolumn{1}{c|}{0}                 & 84.30\%          & \multicolumn{1}{c|}{48.42\%}          & 92.27\%          & 83.21\%          \\ 
\multicolumn{1}{c|}{100}               & 84.31\%          & \multicolumn{1}{c|}{49.02\%}          & 93.13\%          & 87.23\%          \\ 
\multicolumn{1}{c|}{200}               & 82.05\%          & \multicolumn{1}{c|}{47.06\%}          & 92.90\%          & 86.51\%          \\ 
\multicolumn{1}{c|}{300}               & 82.67\%          & \multicolumn{1}{c|}{49.52\%}          & \textbf{93.48\%} & \textbf{88.51\%} \\ 
\multicolumn{1}{c|}{400}               & 84.13\%          & \multicolumn{1}{c|}{48.72\%}          & 92.53\%          & 84.85\%          \\ 
\multicolumn{1}{c|}{500}               & \textbf{84.49\%} & \multicolumn{1}{c|}{\textbf{49.74\%}} & 92.54\%          & 84.82\%          \\ 
\multicolumn{1}{c|}{600}               & 84.10\%          & \multicolumn{1}{c|}{49.03\%}          & 91.75\%          & 78.48\%          \\ \bottomrule
\end{tabular}
\label{tab_offline}
\end{center}
\end{table}

\subsection{Online}\label{sec_online_exp}
According to our ``online" proposed method, the CCSP scores are calculated for each sample after each epoch. Then, the scores are sorted, and $R$ samples with lowest scores are removed only in the following epoch. $R=100$ is used in this part to avoid removing useful samples in addition to the hard ones. In addition, according to the adaptive removability of samples during training, removing 100 samples is sufficient.

Results are shown in Table \ref{tab_online}. Accordingly, this method also improves robustness over the baseline ($R=0$). So, we can conclude that employing online strategy also works well.


\begin{table}[t]
\caption{Clean and robust accuracy of models trained on CIFAR-10 and SVHN datasets using ``online" method.}
  \renewcommand{\arraystretch}{1.1} 
\begin{center}
\begin{tabular}{ccccc}
\toprule
\multirow{3}{*}{$R$} & \multicolumn{4}{c}{Dataset}                                                                    \\ \cline{2-5} 
                                       & \multicolumn{2}{c|}{CIFAR-10}                             & \multicolumn{2}{c}{SVHN}            \\ \cline{2-5} 
                                       & Clean            & \multicolumn{1}{c|}{Robust}           & Clean            & Robust           \\ \hline
\multicolumn{1}{c|}{0}                 & 84.30\%          & \multicolumn{1}{c|}{48.42\%}          & 92.27\%          & 83.21\%          \\ 
\multicolumn{1}{c|}{100}               & 84.03\%          & \multicolumn{1}{c|}{49.66\%}          & 92.78\%          & 88.28\%          \\  \bottomrule
\end{tabular}
\label{tab_online}
\end{center}
\end{table}

\subsection{Mahalanobis Distance}

In addition to the softmax-based methods, the distance between HTS and conditional distribution of the classes can be used to identify them. Two main techniques in this regard are the Mahalanobis distance (MD) \cite{Mahalanobis} and the Relative MD (RMD) \cite{RMD}. These techniques fit a conditional Gaussian distribution $\mathcal{N}(\mu_k, \Sigma)$ to the pre-logit features $h$ for a distribution with $K$ classes. The mean vector and covariance matrix are calculated as:
\begin{equation}
    \mu_k = \frac{1}{N_k} \sum_{i:y_i=k}h_i,
\end{equation}    
\begin{equation}
    \Sigma=\frac{1}{N}\sum_{k=1}^{K}\sum_{i:y_i=k}(h_i-\mu_k)(h_i-\mu_k)^T,
\end{equation}
for $k=1, 2, ..., K$, where $N_k$ is the number of samples in the class with label $k$, and $N$ refers to the number of samples in the dataset. Please note that $\Sigma$ is shared among by all classes. Afterward, the distance of the input $x$ with pre-logits $h_x$ is calculated as:
\begin{equation}
    {MD}_k(h_x) = (h_x - \mu_k)^T\Sigma(h_x-\mu_k),
\end{equation}
\begin{equation}
    {RMD}_k(h_x) = {MD}_k(h_x) - {MD}_0(h_x),
\end{equation}
where ${MD}_0(h_x)$ represents the Mahalanobis distance of $h_x$ to a distribution fitted to the entire training dataset as $\mathcal{N}(\mu_0, \Sigma_0)$. $\mu_0$ and $\Sigma_0$ are calculated as:
\begin{equation}
    \mu_0 = \frac{1}{N} \sum_{i=1}^N h_i,
\end{equation}
\begin{equation}
    \Sigma_0=\frac{1}{N}\sum_{i=1}^N (h_i-\mu_0)(h_i-\mu_0)^T.
\end{equation}
According to these descriptions, we can use Absolute Relative Mahalanobis Distance (ARMD) as an alternative method to CCSP for detecting HTS. ARMD is defined as follows for sample $x$ with label $y$ and pre-logit features $h_x$:
\begin{equation}
    {ARMD(x, y) = |{RMD}_y(h_x)|}.
\end{equation}
By determining the ARMD score for each sample, we can sort the samples in accordance with the calculated scores. After that, we can remove the $R$ samples with the lowest scores to get rid of the hard samples similar to the proposed method with CCSP method. 

The result of this approach can be seen in the Table \ref{tab_maha} where a noticeable improvement over the baseline ($R=0$) can be observed in most cases. As a result, our method is not sensitive to the detection method.

\begin{table}[t]
\caption{Clean and robust accuracy of models trained with the Mahalanobis scoring method instead of CCSP.}
  \renewcommand{\arraystretch}{1.1} 
\begin{center}
\begin{tabular}{ccc}
\toprule
\multirow{2}{*}{$R$}     
                         & \multicolumn{2}{c}{CIFAR-10}                             \\ \cline{2-3} 
                         & Clean            & \multicolumn{1}{c}{Robust}           \\ \hline
\multicolumn{1}{c}{0}   & 84.30\%          & \multicolumn{1}{c}{48.42\%}          \\ 
\multicolumn{1}{c}{100} & 83.77\%          & \multicolumn{1}{c}{49.05\%}          \\ 
\multicolumn{1}{c}{200} & 81.86\%          & \multicolumn{1}{c}{43.94\%}          \\ 
\multicolumn{1}{c}{300} & 84.15\%          & \multicolumn{1}{c}{47.32\%}          \\ 
\multicolumn{1}{c}{400} & 83.64\%          & \multicolumn{1}{c}{48.85\%}          \\ 
\multicolumn{1}{c}{450} & \textbf{84.60\%} & \multicolumn{1}{c}{48.79\%}          \\ 
\multicolumn{1}{c}{500} & 83.77\%          & \multicolumn{1}{c}{\textbf{49.78\%}} \\ 
\multicolumn{1}{c}{550} & 84.38\%          & \multicolumn{1}{c}{49.74\%}          \\ 
\multicolumn{1}{c}{600} & 82.89\%          & \multicolumn{1}{c}{47.75\%}          \\ \bottomrule
\end{tabular}
\label{tab_maha}
\end{center}
\end{table}

\subsection{Clean training instead of removing}
An alternative option for removing the hard samples in our method is using them without any perturbation in training. This can be useful if all the samples in the dataset correlate well with their assigned label.

To investigate the effectiveness of this method, the experiment in section \ref{sec_online_exp} is repeated with clean training of hard samples instead of removing them from dataset. Results are shown in Table \ref{tab_clean}. According to this table, this method is effective in SVHN dataset, but it does not improve the baseline ($R=0$) in CIFAR-10. We believe that the reason for this observation is that the quality of some of the CIFAR-10 images is so substandard that even their clean training can be harmful as can be seen in Fig. \ref{fig_2}.

\begin{table}[t]
\caption{Clean and robust accuracy of models trained on CIFAR-10 and SVHN datasets using ``online" method with clean training of hard samples instead of removing them from dataset.}
  \renewcommand{\arraystretch}{1.1} 
\begin{center}
\begin{tabular}{ccccc}
\toprule
\multirow{3}{*}{$R$} & \multicolumn{4}{c}{Dataset}                                                                    \\ \cline{2-5} 
                                       & \multicolumn{2}{c|}{CIFAR-10}                             & \multicolumn{2}{c}{SVHN}            \\ \cline{2-5} 
                                       & Clean            & \multicolumn{1}{c|}{Robust}           & Clean            & Robust           \\ \hline
\multicolumn{1}{c|}{0}                 & 84.30\%          & \multicolumn{1}{c|}{48.42\%}          & 92.27\%          & 83.21\%          \\ 
\multicolumn{1}{c|}{100}               & 84.58\%          & \multicolumn{1}{c|}{48.44\%}          & 93.73\%          & 88.58\%          \\  \bottomrule
\end{tabular}
\label{tab_clean}
\end{center}
\end{table}

\begin{table}[t]
\caption{Accuracy (\%) of different defense methods on CIFAR-10.}
  \renewcommand{\arraystretch}{1.1} 
\begin{center}
\begin{tabular}{ccc}
\hline
\multirow{2}{*}{Defense}             & \multicolumn{2}{c}{CIFAR-10} \\ \cline{2-3} 
                                     & Clean        & Robust        \\ \hline
\multicolumn{1}{c|}{Standard}        & 84.08        & 48.42        \\
\multicolumn{1}{c|}{MMA}             & 84.40        & 49.20        \\
\multicolumn{1}{c|}{Dynamic}         & 82.97        & 50.27        \\
\multicolumn{1}{c|}{TRADES}          & 82.54        & 51.12        \\
\multicolumn{1}{c|}{Online (R=100)}  & 83.81        & 49.66        \\
\multicolumn{1}{c|}{Offline (R=500)} & 84.49        & 49.74        \\ \hline
\end{tabular}
\label{tab_comp}
\end{center}
\end{table}

\subsection{Are OOD samples learned?}
In this section, we conduct an experiment to show the capability of model in learning the OOD samples during training, which can cause problems in adversarial training. For this purpose, we add 50 random samples from CIFAR-100 dataset to the CIFAR-10 training set.
These samples are randomly chosen from classes that are not shared between CIFAR-10 and CIFAR-100,  
and they are randomly assigned to one of the ten classes in CIFAR-10. So, we can consider these samples as OOD samples in the training set.

Afterwards, the model is trained using the updated dataset for 50 epochs. Results show that model classifies 40 out of these 50 samples correctly in the last epoch of training. In other words, model can learn 80\% of the OOD samples in training, which confirms our claims in this work. As a result, the dataset should be purified before training, as has been done in this study.


\subsection{Comparison with other defenses}

We have also made a comparison with some other recent defenses on CIFAR-10 dataset in Table \ref{tab_comp}.
The defense methods in this table are Dynamic \cite{Dynamic}, TRADES \cite{trades2019}, MMA \cite{MMA}, standard adversarial training \cite{minmax}, and our online and offline proposed methods with the best $R$. Considering the fact that there is a trade-off between clean and robust accuracy \cite{trades2019}, the results show that our method is competitive against other variants of AT. Please note again that our method have a different data-centric approach than those methods.


\section{Conclusion}
Adversarial training is shown to be the most effective existing defense. Therefore, a lot of efforts have been made to improve its result. In this work, we first demonstrated that some of these improvements are made by behaving differently with the near-boundary or hard samples in the training. Accordingly, from a data-centric point of view, we suggested to identify these samples to ignore them during training. For this purpose, we used probablity and distance based methods to detect the hard samples. After that, these samples are removed from the dataset with two  ``offline'' and ``online'' methods. The ``offline'' method removes the hard samples before the training, while the ``online'' removing is done after each epoch in the training. The results on both of the methods show improvements over the baseline.

\end{document}